\documentclass{sage}          

\usepackage{amsthm,amssymb,natbib,url,amsmath}
\usepackage{algorithm}

\usepackage{textcomp}
\usepackage{color}
\usepackage{ifpdf}
\usepackage{url}
\usepackage{enumerate}
\usepackage{array}
\usepackage{flushend}
\usepackage{hyperref}
\usepackage[utf8]{inputenc}

\usepackage{tcolorbox}
\newcommand{\alg}{Algorithm~}
\newtcolorbox[auto counter]{algorithmbox}[2][]{colback=red!5!white,colframe=red!75!black,fonttitle=\bfseries, title=\alg\thetcbcounter: #2,#1}

\journal{Technical Report}
\volume{--}
\issue{--}
\copyrightline{$\copyright$ Claudio Zito}
\firstpage{1}
\lastpage{9}
\doi{--}
\articletype{--}
\pubyear{2020}

\begin{document}
	
\title{Aging Touch: Systematic and Unbiased Presentation of Tactile Stimuli}

\author{Claudio Zito\thanks{Corresponding author; e-mail: C.Zito@bham.ac.uk}
	\thanks{This work is supported by the BBSRC Aging Touch project.}
}


\address{SymonLab, School of Phychology, University of Birmingham, Edgbaston, Birmingham, B15 2TT.}

\maketitle

\begin{abstract}
This report presents the experimental methodology and a step-by-step guide for gathering data on how aging influences tactile surface perception in decision and action. The experiments consist of a set of trials in which the ability to distinguish tactile stimuli is investigated. A robot arm is used to provide a systematic and unbiased presentation of the stimuli.  

\end{abstract}

\section{Introduction}
\label{sec:indroduction}

The aim of this project is to develop a robotic system capable of delivering unbiased tactile stimuli. 
The system is composed of i) a three-DOF parallel robot arm (Force Dimension Delta 3), ii) a six-channel force-torque sensor (NI-DAQ), and iii) a stepper motor (Arduino Uno). The first two components are controlled by a robotic software framework, Golem, which provides control and planning of the robotic arm, as well as the processing of the FT sensor signals (see for further details \cite{bib:zito_2016, bib:zito_2019, bib:zito_2013, bib:zito_w2013, bib:zito_2012, bib:rosales_2018, bib:zito_ecvp_2012, bib:zito_w2012, bib:zitoWRRS2019a, bib:zitoWRRS2019b, zito_2020, heiwoltICORR2019, bib:stuberICRA2018, bib:zitoWRRS2019b}).

\section{Hardware setup}
\label{hardware_setup}

In order to execute the touch demo the following steps need to be done before running the code:
\begin{enumerate}
    \item Switch on the PC and log in into the SymonLab account.
    
    \item Connect the Delta:
    \begin{enumerate}
        \item Check that the Delta 3 is connected to the PC via an ethernet cable and 
        \item switch on the robot controller (e.g. Force Dimension box).
    \end{enumerate}
    
    \item Connect the FT sensor:
    \begin{enumerate}
        \item Check that the NI-DAQ is connected to the PC via an ethernet cable and
        \item switch on the NI box.
    \end{enumerate}
    
    \item Connect the stepper motor:
    \begin{enumerate}
        \item connect the Arduino board to the PC with the USB cable,
        \item turn on the stepper motor box and
        \item switch on the controller (small black buttun on the front of the box)
    \end{enumerate}
    
    \item On the PC run the following steps to initialise the Delta
    \begin{enumerate}
        \item Open the HapticInit program on the desktop.
        \item Click on the initialisation button and wait for the initialisation to finish.
        \item The same button should now show ``check'' instead of initialisation. Click it and wait for the robot to finish the check.
        \item Close the hapticInit program. If you don't no other program can connect to the same device.
        \item Run the center.exe program on the desktop. This program re-run the calibration if needed. Terminate the program as soon as the calibration is over by selecting the console and press ``q'', or closing the console by the ``x'' button.
    \end{enumerate}
\end{enumerate}

\section{Golem framework}
\label{sec:golem}

Golem is an open-source, multi-platform framework for control and planning of robotic systems. This section presents a detailed overview of the framework.

\subsection{Major dependencies}
\label{sec:golem_dependencies}

The framework requires the following dependencies:

\begin{itemize}
    \item Expat: open-source library for processing xml files.
    \item Freeglut: open-source library for computer graphics.
    \item Boost: free peer-reviewed portable C++ source libraries. Required version $>=$ 1.58.
    \item OpenCV 3.4.X: open-source computer vision library.
    \item PCL: open-source point cloud library. Required version $>=$ 1.7.
    \item Force Dimension Haptic SDK: software interface for all Force Dimension products.
    \item National Instrument DAQ: data acquisition software for NI force sensors.
\end{itemize}

\subsection{Installation}
\label{sec:golem_installation}

For a detailed example of how to install the framework on Ubunto 16.04 LTS see the README.md file in the main folder where the software is installed, see Sec.~\ref{sec:golem_folder_structure}.

\subsection{Folder Structure}
\label{sec:golem_folder_structure}

The chosen folder in which you have chosen to setup the framework looks like this:

\begin{verbatim}
Projects\
   bin\
   Golem.RobotLab\
   lib\
Software\
\end{verbatim}

In the SymonLab's PC this folder is
\begin{verbatim}
D:\Programming\Development\
\end{verbatim}

Projects contains: i) the cloned repository of the Golem code, ii) bin which contains the compiled programs, and iii) lib which contains the Golem compiled libraries. Software containes the dependensies as described in Sec.~\ref{sec:golem_dependencies}.

The main directory of the framework (Golem.RobotLab) has the following structure:

\begin{verbatim}
BUILD.vcXX.x64\
cmake\
docs\
packages\
CmakeLists.txt
Copyright.txt
Doxyfile
License.txt
README.md
Readme.txt
\end{verbatim}

The folder Build.vcXX.x64 contains the Visual studio projects and solution built by the CMake program. The label vcXX identifies the version of Visual Studio (i.e. target compiler). The label x64 specifies that the target machine, i.e. x64 target a 64-bit machine while x86 a 32-bit machine.  

The cmake folder contains utilities files for the CMake compilation. CMake is a cross-platform free and open-source software tool for managing the build process of software using a compiler-independent method. The build process with CMake takes place in two stages. First, standard build files are created from configuration files (i.e. the files that go into the Build folder). Then the platform's native build tools (e.g. Visual Studio's compiler) are used for the actual building. The compiled programs (.exe) are stored in the bin folder, while the compiled library (.lib) are stored in the lib folder.

\subsection{AppSymons}
\label{sec:golem_appsymons}

AppSymons is the main program for running the aging touch experiments. From the command window the program can be launched by typing the following commands:

\begin{verbatim}
C:\>D:
D:\>cd D:\Programming\Development\Projects\bin
<bin_folder>\>GolemAppSymons.exe GolemAppSymons_RobotDelta3SM.xml
\end{verbatim}

The first two commands move you to the bin folder where the program is stored. The last command launch the program with its configuration file.

The programs opens a windows with the simulation of the robot, and a console. Note: the console is only for the outputs of the program. To interact with the program the window showing the robot must be the active window (i.e. selected by a click of the left button of the mouse).

By typing ``?'' (question mark) the ''help`` command of the menu is visualised, printing on the console a list of available commands. There are no buttons in the interface but you can interact via keyboard. By pressing ``R'' (capital r), the ``run'' menu is loaded. To run the demo just press ''D`` (capital d).

Via ``Z'' (capital z) you access a set of few utilities to control the delta robot, read the FT signals, and visualise the data collected in an experiment.

\subsection{Demo}
\label{sec:golem_demo}

The demo is run by sequentially pressing the ``R'' and ``D'' key. At this point, the demo is loaded and ready to be executed. The operator can exit the demo at any time by simply press the <esc> key. 

First the demo require the user to select if the demo has to run in debug mode. 
\begin{verbatim}
Debug mode: YES (Continue Y/N)
\end{verbatim}
Press ``Y'' (capital y) for activating the debug mode, and ``N'' (capital n) otherwise.
The debug mode stops the demo in critical points, called hereafter break-points, and requires an input from the user before continuing with the program. Additionally, the stepper motor is not integrated in this framework, so the user needs to manually command the stepper motor for moving the two pins. The debug mode allows the operator to insert the distance required in the Arduino interface while the main demo program is waiting. 

The following step are executed only at the beginning of the data collection. It collects the data of the participant.
\begin{verbatim}
Enter unique ID: 
Enter participant name: 
Enter participant surname: 
Enter participant age: 
Gender: Female
Enter participant notes: 
\end{verbatim}
Each question pops up after the previous has been completed and <enter> pressed. The fifth question has only two option: FEMALE or MALE. Manual insertion by typing the gender is not allowed, but the correct gender can be selected by pressing the <tab> key. Finally, the last entry is optional.

Next the robot needs to be moved to a initial pose. This is the home pose for the robot. Between each stage the robot will go back to this pose. The pose was empirically chosen to be around 2 cm above the participant's finger.
\begin{verbatim}
[DEMO]: moving to init pose (Continue Y/N)
\end{verbatim}
Before performing this action, the program presents its first break-point for safety reason. The operator must press ``Y'' (capital y), otherwise the program will wait. IF ``N'' is pressed the demo is cancelled.

After the robot Delta has been moved to the initial pose the following error could happen, cancelling the demo.
\begin{verbatim}
No poses for the stepper motor
\end{verbatim}
This error can only happen if the user has not define a set of distances for the stepper motor in the config file (see Sec.~\ref{sec:golem_config_file}).

The next message is just informative for the operator and does not require any input.
\begin{verbatim}
[DEMO] Training 1/10 presentation: Single Pin First
\end{verbatim}
This type of message is printed at each trial. The message first shows the label of this trial (Training or Trial), and identifies if the current trial is used for training the participant or as a real trial. The next information is the number of the trial out of the total number of trials. Note: the number of the trial is reset to 1 after the training is completed. The last information is about the type of presentation (Single Pin First or Two Pins First). 

The next message is again just informative. Each distance for the stepper motor is randomly sampled in an uniform manner at each trial. However, the program guaranties a balanced number of presentations for each distance. The print shows how many times a distance has been presented and if this distance is still available for future trials.

At each trial, even for the training, the program will present a second break-point. The program asks the operator to manually move the stepper motor to the selected distance before starting the trial.
\begin{verbatim}
[DEMO]: move stepper motor to 1.0 [mm] 363.00 [step]
\end{verbatim}
In this case, the program has selected 1 mm for the two pins presentation and the operator must input 363.00 in the input bar on the Arduino interface.

The next messages are only for information. The robot will move left or right in the horizontal plane to align the single pin or the two pins with the participant finger. Then it will move downward to create a contact on the finger's tip. The two-part motion is required for not biasing the participant, since both presentation are presented from a vertical movement. The robot arm will stop as soon as a contact is perceived by the FT sensor and collects the FT value for a pre-defined number of times (see Sec.~\ref{sec:golem_config_file}). The collected forces are shown on the console, with the following format:
\begin{verbatim}
FT [touched=TRUE] [timestamp] [fx] [fy] [fz] [tx] [ty] [tz]
\end{verbatim}
The touched flag shows if the robot has stopped (and the data collection started) after a contact or at the end of the robot motion. If touched is FALSE the forces collected may be meaningless, because the robot has not reached the participant's finger.

After the first presentation, the robot automatically will move to the next presentation and execute the same steps. At the end of the second presentation the program will present a third and final break-point to collect the participant's response.
\begin{verbatim}
[Demo] Response: First
\end{verbatim}
The participant must communicate the response and the operator inputs the response in the system. Typing is not allowed, but with the <tab> key the operator can switch between the two options and select the correct one by pressing <enter>. The options are the following: ``First'', as in the two pins have been used in the first presentation, or ``Second'', if the participant is convinced that the two pins were used in the second poke.

The program outputs on the console if the answer is correct or wrong as information only to the operator. 

At the end of the trial, the data are temporary saved in this location
\begin{verbatim}
<bin_folder>\data\<participant_id>\tmp.csv
\end{verbatim}
At the end of the experiment all the data will be saved in the same folder, as
\begin{verbatim}
<bin_folder>\data\<participant_id>\data.xml
\end{verbatim}
The file data.xml contains the links to two .csv files. The first is automatically named as
\begin{verbatim}
data-<participant_id>-<participant_surname>.csv 
\end{verbatim}
and contains all the personal data of the subject. The second is automatically named 
\begin{verbatim}
data-<participant\_id>-trial.csv 
\end{verbatim}
and contains all the trial data. See Sec.~\ref{sec:golem_data} for a detailed explanation on the .csv data format.

\subsection{The config file}
\label{sec:golem_config_file}

The configuration file is the xml file specified as the first and only argument when launching the program, see Sec.~\ref{sec:golem_appsymons}.

The configuration file is quite complex but it is divided in modules. The parts that affect the experiment is denoted as ``Demo''.
\begin{verbatim}
 <demo data_name="data.demo">
    <wpose name="rel_poking_single_pin" v1="0.0" v2="-0.01" v3="-0.05"/>
    <wpose name="rel_release_single_pin" v1="0.0" v2="0.0" v3="+0.05"/>
    <wpose name="rel_poking_two_pins" v1="0.0" v2="+0.01" v3="-0.05"/>
    <wpose name="rel_release_two_pins" v1="0.0" v2="0.0" v3="+0.05"/>
    <wpose name="gbl_zero" v1="0.0" v2="0.0" v3="0.0"/>

    <!--Two pins conditions: distances-->
    <!--For young fellas-->
    <smpose name="sm_commands" dim="2" c1="0.0001" c2="0.0"/>
    <smpose name="sm_commands" dim="2" c1="0.0003" c2="0.0"/>
    <smpose name="sm_commands" dim="2" c1="0.0006" c2="0.0"/>
    <smpose name="sm_commands" dim="2" c1="0.001" c2="0.0"/>
    <smpose name="sm_commands" dim="2" c1="0.0013" c2="0.0"/>
    <smpose name="sm_commands" dim="2" c1="0.0016" c2="0.0"/>
    <smpose name="sm_commands" dim="2" c1="0.002" c2="0.0"/>

    <!--For older fellas-->
    <!--<smpose name="sm_commands" dim="2" c1="0.001" c2="0.0"/>
    <smpose name="sm_commands" dim="2" c1="0.0013" c2="0.0"/>
    <smpose name="sm_commands" dim="2" c1="0.0016" c2="0.0"/>
    <smpose name="sm_commands" dim="2" c1="0.002" c2="0.0"/>
    <smpose name="sm_commands" dim="2" c1="0.0024" c2="0.0"/>
    <smpose name="sm_commands" dim="2" c1="0.003" c2="0.0"/>
    <smpose name="sm_commands" dim="2" c1="0.0036" c2="0.0"/>
    <smpose name="sm_commands" dim="2" c1="0.0043" c2="0.0"/>
    <smpose name="sm_commands" dim="2" c1="0.005" c2="0.0"/>
    <smpose name="sm_commands" dim="2" c1="0.006" c2="0.0"/>-->

    <touch sensor="FTDAQ+FTDAQ_Delta3" event_time_wait="0.10"
    movement_duration="2.0" poking_duration="5.0">
      <threshold v1="0.5" v2="0.5" v3="0.25" w1="0.1" w2="0.1" w3="0.1"/>
      <motion_single_pin v1="0.0" v2="+0.024" v3="0.0"/>
      <motion_two_pins v1="0.0" v2="-0.018" v3="0.0"/>
      <poking v1="0.0" v2="0.0" v3="-0.02"/>
      <init v1="-0.1" v2="0.0" v3="-0.075"/>
    </touch>

    <experiment_data participant_ext_file=".csv" trial_ext_file=".csv" 
    number_training_trials="1" training_index="1" number_presentations="10"
    number_ftdata_recordings="10"  path="./data/"></experiment_data>
  </demo>
\end{verbatim}

The wposes are a set of pre-defined workspace coordinates in 3D. Each pose is labelled with a name. Multiple poses can have the same label. From the GolemAppSymons it is possible to move the delta robot to these poses by pressing ``Z'' and ``R'' which interprets the pose as a relative pose, or by pressing ``Z'' and ``G'' which interprets the pose as a global pose. There fore if our selected pose is $<0,0,0>$, the ``ZR'' option will not move the delta (i.e. the target position becomes the current pose of the delta plus zero), while the ``ZG'' option will send the robot delta to the origin.

The smposes define the command for the stepper motor (sm). The commands are in meters and the variable which controls the distance of the stepper motor is c1. The first set of distances are for people younger than 35 years old. A second identical configuration file is available in the bin folder as 
\begin{verbatim}
GolemAppSymons_RobotDelta3Sm_elderly.xml 
\end{verbatim}
which has the second set of distances uncommented instead of the first.

The section ``touch'' controls the pre-defined movements of the robot during the demo, as well identified the FT sensor to be used (sensor=``FTDAQ+FTDAQ\_Delta3''). The rest of the parameters have the following meaning:
\begin{itemize}
    \item event\_time\_wait The time of waiting between each FT readings.
    \item movement\_duration The minimum time for free space movements, when the robot is not assume to get in contact with the finger.
    \item poking\_duration The minimum time to move the delta in a downward motion towards the finger.
    \item threshold The threasolds for detecting a contact and stop the delta poking motion. Six thresholds for the six channels of the FT sensor.
    \item motion\_single\_pin Relative motion of the delta to move the single pin just above the finger (i.e. 24 cm on the left from the home pose of the robot).
    \item motion\_two\_pins Relative motion of the delta to move the two pins just above the finger (i.e. 18 cm on the right from the home pose of the robot). This motion is also adjusted in the program with an offset to ensure that the two pins hit the center of the finger. The offset is computed as $- 0.5 * distance$, and distance is the c1 variable of the selected smpose.
    \item poking Relative motion to reach and touch the finger (downward movement of 2 cm).
    \item init Home pose. Global pose for the robot to move to before starting the data collection.
\end{itemize}

The experiment\_data section contains the parameters for the data collection, as follows:
\begin{itemize}
    \item participant\_ext\_file The file extension used to save the participant data.
    \item trial\_ext\_file The file extension used to save the trial data.
    \item num\_training\_trials The number of training trials before starting collecting the real data for the experiments.
    \item training\_index The fixed index of the training trail in which the maximum distance is presented to the participant.
    \item num\_presentations The number of times each distance for the two pins (smpose) must be presented to the participant. The total number of trials is composed by the number of training plus the number of presentation times the number of distances in the smpose.
    \item num\_ftdata\_recordings The number of FT data recorded when in contact with the fingers. Note: event\_time\_wait defines the frequency of this recordings.
    \item path The path in which the data will be saved.
\end{itemize}

\subsection{Data}
\label{sec:golem_data}

The data is saved in a data.xml file. If the participant id is ``dfs'' and the surname is ``foo'', the file will look like this
\begin{verbatim}
<?xml version="1.0" encoding="utf-8"?>

<golem>
  <data data_name="data.demo">
    <participant ext_file=".csv" filename="dfs-foo.csv"></participant>
    <trials ext_file=".csv" filename="dfs-foo-trial.csv"></trials>
  </data>
</golem>
\end{verbatim}

The file data-dfs-foo.csv has only one line with the following structure
\begin{verbatim}
[ID] [NAME] [SURNAME] [AGE] [GENDER] [NOTES]
\end{verbatim}

The file data-dfs-foo-trial.csv has a line per trial with the following structure
\begin{verbatim}
<[No OF TRIALS]> [ID] [No] [PRESENTATION] [FTDATA FOR FIRST PRESENTATION] 
[FTDATA FOR SECOND PRESENTATION] [DISTANCE] [RESPONSE]
\end{verbatim}
Note: [No OF TRIALS] is only written once at the beginning.
Each FT Data is saved as
\begin{verbatim}
<[No OF READINGS]> [TIME STAMP] [V.X] [V.Y] [V.Z] [W.X] [W.Y] [W.Z]
\end{verbatim}

\section*{Supplementary Material}

For a full documentation of the Golem API open the file
\begin{verbatim}
<your path>/Projects/Golem.RobotsLab/docs/html/index.html
\end{verbatim}

\bibliographystyle{abbrv}
\bibliography{refs}

\begin{thebibliography}{10}

\bibitem{bib:zito_ecvp_2012}
M.~Di~Luca, T.~E. Vivian-Griffiths, J.~L. Wyatt, and C.~Zito.
\newblock Grasping a shape with uncertain location.
\newblock In {\em Perception}, volume~41, page 253, 2012.

\bibitem{heiwoltICORR2019}
K.~Heiwolt, C.~Zito, M.~Nowak, C.~Castellini, and R.~Stolkin.
\newblock Automatic detection of myocontrol failures based upon situational
  context information.
\newblock In {\em Proceeding of {IEEE/RAS-EMBS} International Conference on
  Rehabilitation Robotics ({ICORR})}, 2019.

\bibitem{bib:rosales_2018}
C.~J. Rosales, F.~Spinelli, M.~Gabiccini, C.~Zito, and J.~L. Wyatt.
\newblock Gpatlasrrt: a local tactile exploration planner for recovering the
  shape of novel objects.
\newblock {\em International Journal of Humanoid Robotics, Special Issue
  'Tactile perception for manipulation: new progress and challenges'}, 15,
  2018.

\bibitem{bib:stuberICRA2018}
J.~Stüber, M.~Kopicki, and C.~Zito.
\newblock Feature-based transfer learning for robotic push manipulation.
\newblock In {\em Proceeding of {IEEE} International Conference on Robotics and
  Automation ({ICRA})}, 2018.

\bibitem{bib:zito_2016}
C.~Zito.
\newblock {\em Planning simultaneous perception and manipulation}.
\newblock PhD thesis, University of Birmingham, 2016.

\bibitem{zito_2020}
C.~Zito.
\newblock Thesis proposal.
\newblock {\em CoRR arXiv preprint, arXiv:2002.03306 [cs.RO]}, 2020.

\bibitem{bib:zitoWRRS2019b}
C.~Zito, M.~Adjigble, B.~Denoun, L.~Jamone, M.~Hansard, and R.~Stolkin.
\newblock Metrics and benchmarks for remote shared controllers in industrial
  applications.
\newblock In {\em Proc. of the Workshop on Task-Informed Grasping (TIG-II):
  From Perception to Physical Interaction, Robotics: Science and Systems
  ({RSS})}, 2019.

\bibitem{bib:zitoWRRS2019a}
C.~Zito, T.~Deregowski, and R.~Stolkin.
\newblock 2d linear time-variant controller for human's intention detection for
  reach-to-grasp trajectories in novel scenes.
\newblock In {\em Proc. of the Workshop on Task-Informed Grasping (TIG-II):
  From Perception to Physical Interaction, Robotics: Science and Systems
  ({RSS})}, 2019.

\bibitem{bib:zito_w2013}
C.~Zito, M.~Kopicki, R.~Stolkin, C.~Borst, F.~Schmidt, M.~A. Roa, and J.~Wyatt.
\newblock Sequential re-planning for dextrous grasping under object-pose
  uncertainty.
\newblock In {\em Workshop on Manipulation with Uncertain Models, Robotics:
  Science and Systems ({RSS})}, 2013.

\bibitem{bib:zito_2013}
C.~Zito, M.~Kopicki, R.~Stolkin, C.~Borst, F.~Schmidt, M.~A. Roa, and J.~Wyatt.
\newblock Sequential trajectory re-planning with tactile information gain for
  dextrous grasping under object-pose uncertainty.
\newblock In {\em {IEEE} Proc. Intelligent Robots and Systems ({IROS})}, 2013.

\bibitem{bib:zito_2019}
C.~Zito, V.~Ortienzi, M.~Adjigble, M.~S. Kopicki, R.~Stolkin, and J.~L. Wyatt.
\newblock Hypothesis-based belief planning for dexterous grasping.
\newblock {\em CoRR arXiv preprint, arXiv:1903.05517 [cs.RO] (cs.AI)}, 2019.

\bibitem{bib:zito_w2012}
C.~Zito, R.~Stolkin, M.~Kopicki, M.~Di~Luca, and J.~Wyatt.
\newblock Exploratory reach-to-grasp trajectories for uncertain object poses.
\newblock In {\em Workshop of Beyond Robot Grasping, {IEEE/RSJ} Intelligent
  Robots and Systems ({IROS})}, 2012.

\bibitem{bib:zito_2012}
C.~Zito, R.~Stolkin, M.~S. Kopicki, and J.~L. Wyatt.
\newblock Two-level rrt planner for robotic push manipulation.
\newblock In {\em {IEEE} Proc. Intelligent Robots and Systems ({IROS})}, pages
  678--685, 2012.

\end{thebibliography}
\end{document}